\documentclass[lettersize,journal]{IEEEtran}

\usepackage{amsmath,amsfonts}
\usepackage{algorithmic}
\usepackage{algorithm}
\usepackage{array}
\usepackage[caption=false,font=normalsize,labelfont=sf,textfont=sf]{subfig}
\usepackage{textcomp}
\usepackage{stfloats}
\usepackage{url}
\usepackage{verbatim}
\usepackage{graphicx}
\usepackage{cite}

\usepackage{bm}
\usepackage{booktabs}

\newtheorem{theorem}{Theorem}

\graphicspath{{figures/}}

\begin{document}

\title{Multi-Robot Pursuit in Parameterized Formation via Imitation Learning}
\author{
    Jinyong Chen, Rui Zhou, Zhaozong Wang, Yunjie Zhang, and Guibin Sun
    \thanks{This work was supported in part by the STI 2030-Major Projects under Grant 2022ZD0208804, and in part by the National Natural Science Foundation of China under Grant 62473017, and in part by the China Postdoctoral Science Foundation under Grant 2023M740185. \emph{(Corresponding author: Guibin Sun)}}
    \thanks{J.~Chen, R.~Zhou, Z.~Wang, Y.~Zhang and G.~Sun are with the School of Automation Science and Electrical Engineering, Beihang University, Beijing, BJ 100083 China (e-mail: chenjinyong@buaa.edu.cn; zhr@buaa.edu.cn; wangzz0925@buaa.edu.cn; zhangyunjie@buaa.edu.cn; sunguibinx@buaa.edu.cn).}
}



\maketitle

\begin{abstract}
    This paper studies the problem of multi-robot pursuit of how to coordinate a group of defending robots to capture a faster attacker before it enters a protected area. 
    Such operation for defending robots is challenging due to the unknown avoidance strategy and higher speed of the attacker, coupled with the limited communication capabilities of defenders. 
    To solve this problem, we propose a parameterized formation controller that allows defending robots to adapt their formation shape using five adjustable parameters. 
    Moreover, we develop an imitation-learning based approach integrated with model predictive control to optimize these shape parameters.
    We make full use of these two techniques to enhance the capture capabilities of defending robots through ongoing training. 
    Both simulation and experiment are provided to verify
    the effectiveness and robustness of our proposed controller. 
    Simulation results show that defending robots can rapidly learn an effective strategy for capturing the attacker, and moreover the learned strategy remains effective across varying numbers of defenders. 
    Experiment results on real robot platforms further validated these findings. 
\end{abstract}


\section{Introduction}
\IEEEPARstart{P}{URSUIT-EVASION} games have been an attractive and widely studied topic for decades in the realm of multi-robot systems~\cite{lopez2019solutions,zhang2022tcyber,mrios2024tnnls} because of their potential applications in maneuvering target tracking~\cite{mukherjee2022tcyber}, search and rescue operations\cite{oyler2016pursuit}, and intelligent transportation~\cite{wan2021improved}, etc. 
Compared to the single-purser single-evader case~\cite{teng2014self}, multi-robot pursuit is a much more challenging problem that contains cooperation among partners and competition with enemies~\cite{zhang2022game}. 
The existing literature on multi-robot pursuit includes both non-learning and learning methods.

Non-learning methods can be organized into two categories: deterministic and heuristic~\cite{de2021decentralized}. 
Deterministic methods rely on conventional mathematical tools to address the problem. 
In these methods, the capture condition can be obtained by solving the differential equations describing the trajectory of the purser~\cite{shneydor1998missile}, or the system is formulated as a game where pursers optimize an objective function~\cite{pan2023tsmc}. 
The inadequacy of these methods lies in their suitability solely for simplistic scenarios with idealistic assumptions. 
The derivation of closed-form analytical solutions or formulation of an appropriate objective function is arduous, especially when dealing with numerous robots or motion constraints.
Inspired by behavior-based approaches, heuristic methods~\cite{muro2011wolf, angelani2012collective, janosov2017group} utilize computational simulations to identify emergent behavior from local observations. 
Angelani~\cite{angelani2012collective} employs Vicsek-like models to simulate two groups of self-propelled organisms and characterize many aspects of the predation phenomenon. 
In~\cite{muro2011wolf}, the simulation experiments conducted by the authors illustrate that the behavior witnessed during wolf-pack hunting can be emulated using basic rules. 
Janosov et al.~\cite{janosov2017group} introduce more realistic constraints to validate the Vicsek model. However, these methods require the positions of all companions and the velocity or maneuvering strategy of the evader to obtain the optimal pursuit strategy, which is not always available in practice. 
Due to the challenges of creating general multi-robot pursuit methods under partial observation and practical constraints, researchers have increasingly relied on learning methods to develop pursuit strategies tailored to specific scenarios

Learning methods contiguously improve the pursuit strategy by utilizing data generated during interactions with the environment. 
Due to its well-established theoretical foundation~\cite{schulman2017proximal, lillicrap2015continuous} and the impressive fitting ability of deep neural networks, reinforcement learning has emerged as a popular approach for tacking pursuit-evasion problems~\cite{desouky2011q, awheda2015residual, kokolakis2022safety}. 
These works~\cite{desouky2011q, awheda2015residual, kokolakis2022safety} formulate the pursuit-evasion problem as a differential game and use RL algorithms to learn the optimal policy. 
Among them, the algorithm proposed in~\cite{kokolakis2022safety} achieves the finite-time capture of the evader in an unknown environment.
However, traditional RL algorithms may perform poorly when extended to multi-robot pursuit because the inherent non-stationarity and variance increase as the number of robots grows~\cite{lowe2017multi}. 
To this end, Lowe et al.~\cite{lowe2017multi} introduce a widely recognized multi-robot reinforcement learning algorithm, commonly referred to as the multi-robot deep deterministic policy gradient (MADDPG) algorithm.
The MADDPG algorithm employs an adaptation of actor-critic methods that consider action policies of other robots and is capable of learning policies in mixed cooperative-competitive environments. 
Thus, this algorithm and its various variants~\cite{pesce2020improving, kim2019message, li2019robust} have been widely used to address the multi-robot pursuit problem. 
Wan et al.~\cite{wan2021improved} build a control-oriented framework to implement multi-robot pursuit in an environment with obstacles. 
Zhang et al.~\cite{zhang2022game} present a vectorized extension of MADDPG with a target prediction network to coordinate a group of unmanned aerial vehicles (UAVs) to capture a moving target in the urban environment.
More articles~\cite{wang2020cooperative, bilgin2015approach, wu2022crafting, ye2022pursuit} using RL to study the multi-robot pursuit problem can be found in the existing literature.

These existing studies have explored various approaches to applying learning algorithms in multi-robot pursuit scenarios. However, most of these assume that each robot can access the information of all other robots without considering distance limitations, which may not hold in distributed systems with partial perspectives.
Moreover, these RL-based methods rely on trial and error without leveraging existing knowledge of cooperative control to enhance robot cooperation. This results in a high frequency of risky behaviors during the learning process, which may be unacceptable for practical applications with stringent security requirements. 
Introducing cooperative control techniques in multi-robot pursuit is considered a promising direction to simplify the complexity and improve safety. For example, robots move in a formation with variable shapes to capture opponents\cite{chipade2019herding, chipade2021multiagent}. Nevertheless, in these studies, robots are unable to improve pursuit strategies through learning.

Motivated by this, this paper explores leveraging both cooperative control knowledge and learning strategies to enhance multi-robot pursuit. The basic idea is to achieve internal cooperation using a parameterized formation controller that flexibly manipulates the shape of position distribution through five shape parameters, thereby formulating the multi-robot pursuit problem as a decision-making problem involving these parameters. In this context, an imitation learning approach incorporating model predictive control is proposed to continuously improve the pursuit strategy in interaction with the environment.

The main contribution of this paper is three-fold. 
First, it provides a scheme that combines cooperative control and learning algorithms to enhance multi-robot pursuit. By introducing a parameterized formation controller, the multi-robot pursuit problem can be transformed into a simpler online decision-making problem with shape parameters. 
Second, it presents an imitation-learning based approach integrating model predictive control to enable multi-robot autonomous pursuit. This integration reduces the need for real samples and lowers trial-and-error costs. Moreover, the learned strategy shows strong generalization across varying numbers of robots. 
Third, it validates the proposed scheme to a real robot-swarm system. 
The experiment demonstrates the successful adaptation of the learned strategy in the challenging real-world applications. This is to benefit from the fusion of parameterized formation controller and improved imitation-learning approach. 

The remainder of this paper is organized as follows. Section~\ref{Sec_preliminary} introduces the preliminaries and formulates the multi-robot pursuit problem. 
Section~\ref{Sec_method} presents the proposed method in detail.
Simulation and experiment results are shown in Section~\ref{Sec_simulations} and~\ref{Sec_experiment}, respectively. Section~\ref{Sec_conclusion} gives the conclusion of the paper.

\section{Problem Formulation}
\label{Sec_preliminary}

\begin{figure}
    \centering
    \includegraphics[width=0.8\linewidth]{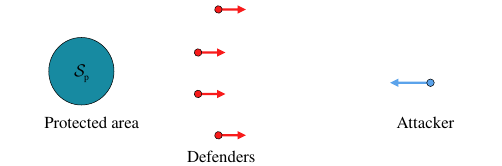}
    \caption{The graphical illustration of the problem setting. 
    }
    \label{Fig_problem_formulation}
\end{figure}

Consider a protected area with center $p_{\rm p}$ and a radius of $\rho_{\rm p}$, denoted as $\mathcal{S}_{\rm p}=\{ p\in \mathbb{R}^2:\Vert p-p_{\rm p} \Vert \leq \rho_{\rm p} \}$, and an attacker who poses a threat to the protected area. The main task of this paper is to formulate a strategy for a group of defenders to capture the attacker before it enters the protected area. A graphical illustration is given in Fig.~\ref{Fig_problem_formulation}. 

Defenders can exchange local information with neighbors within their communication range $r_{\rm com}$ to enable internal cooperation and access the state the attacker to take corresponding actions. The topology of the defenders can be described using an undirected graph $\mathcal{G} = \{ \mathcal{V}, \mathcal{E} \}$, where the vertex set $\mathcal{V} = \{1,2,\dots, n \}$ represents the $n$ defenders, and the edge set $\mathcal{E} = \{ (i,j): r_{ij} < r_{\rm com}, i,j\in\mathcal{V} \}$ with $r_{ij}=\Vert p_{i}-p_{j}\Vert$, represents the existing communication connections. Thus, the neighbor set if $i$ is $\mathcal{N}_i = \{ j\in\mathcal{V}:(i,j)\in \mathcal{E} \}$. Here, the operator $\Vert \cdot \Vert$ represents the Euclidean norm of a vector.

Assume that the dynamics of both defenders and the attacker can be represented by a damped second-order integrator:
\begin{equation*}
    \begin{cases}
        \dot{p}_i = v_i\\
        \dot{v}_i = u_i - C_{\rm d}v_i
    \end{cases} ,\ i=1,\dots,n
    \label{Equ_dynamics_defender}
\end{equation*}
and
\begin{equation*}
    \begin{cases}
        \dot{p}_{\rm a} = v_{\rm a} \\
        \dot{v}_{\rm a} = u_{\rm a} - C_{\rm a}v_{\rm a}
    \end{cases}
    \label{Equ_dynamics_attacker}
\end{equation*}
where $p\in \mathbb{R}^2$, $v\in \mathbb{R}^2$, and $u\in \mathbb{R}^2$ represent the position, velocity, and control input, respectively; $C_{\rm d}$ and $C_{\rm a}$ are the drag coefficients of the defender and attacker.

The attacker can avoid defenders while approaching the protected area, and its strategy is expressed as
\begin{equation}
    u_{\rm a} = k_{\rm ap}\dfrac{p_{\rm a}-p_{\rm p}}{\Vert p_{\rm a}-p_{\rm p}\Vert} + k_{\rm ad}\dfrac{p_{\rm a}-p_{\rm d}^{'}}{\Vert p_{\rm a}-p_{\rm d}^{'}\Vert}g(\Vert p_{\rm a}-p_{\rm d}^{'}\Vert,r_{\rm ad}^{\rm safe},r_{\rm ad}^{\rm avo})
    \label{Equ_control_attacker}
\end{equation}
with 
\begin{equation*}
    g(r,r_1,r_2) = \begin{cases}
        \frac{1}{r-r_1} \left[1+\cos\left(\pi\frac{r-r_1}{r_2-r_1}\right)\right], & r_1< r\leq r_2 \vspace{3pt} \\
        0,                                                                      & r>r_2
    \end{cases}
    \label{Equ_weight}
\end{equation*}
where $r_{\rm ad}^{\rm safe}$ is the safe distance, and $r_{\rm ad}^{\rm avo}$ is the distance threshold at which the attacker begins to avoid defenders; $g(r,r_1,r_2)$ is a monotonically decreasing function mapping $(r_1,+\infty)$ to $[0,+\infty)$.

The maximum input amplitudes of the defender and attacker are denoted as $u_{\rm d}^{\rm max}$ and $u_{\rm a}^{\rm max}$, namely $\Vert u_i\Vert \leq u_{\rm d}^{\rm max}$ and $\Vert u_{\rm a}\Vert \leq u_{\rm a}^{\rm max}$. It is assumed that $u_{\rm d}^{\rm max} < u_{\rm a}^{\rm max}$ and $C_{\rm d} > C_{\rm a}$. Consequently, the attacker has superior maneuverability and velocity compared to the defenders, as $v_{\rm d}^{\rm max}=\frac{u_{\rm d}^{\rm max}}{C_{\rm d}} < \frac{u_{\rm a}^{\rm max}}{C_{\rm a}}=v_{\rm a}^{\rm max}$. In this context, a single defender lacks the capability to capture the attacker, necessitating cooperation among multiple defenders.

\section{Methodology}
\label{Sec_method}

This section details the principles of the proposed method, including a parameterized formation representation leveraging cooperative control, a negotiation strategy for distributed implementation, and an imitation learning approach combined with model prediction.

\subsection{Multi-robot Pursuit in Parameterized Formation}

\begin{figure}
    \centering
    \includegraphics[width=0.8\linewidth]{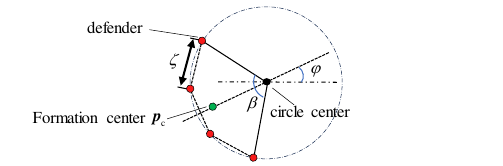}
    \caption{Parameterized formations representation.}
    \label{Fig_parametric_formation}
\end{figure}

To leverage cooperative control theory to facilitate multi-robot pursuit, defenders move in an arc formation parameterized by a shape parameter vector $\theta=[p_{\rm c}^\top, \varphi, \zeta, \beta]$.
The reference positions in the formation pattern, denoted as $\mathcal{P}=\{p_i^{*}\}_n$, are uniquely determined by the shape parameter $\theta$ and robot number $n$, that is, $\mathcal{P}=\mathcal{F}(\theta,n)$. The specific expression is as follows:
\begin{equation*}
    p_i^{*} = p_{\rm c} + q_i - \frac{1}{n}\sum_{k=1}^{n}q_k,\ i=1,2,\dots,n
\end{equation*}
with 
\begin{equation*}
    q_i=
    \begin{cases}
        0, & i=1\\
        q_{i-1} + \zeta[\cos{\psi_i}, \cos{\psi_i}]^\top, & i=2,3,\dots, n
    \end{cases}
\end{equation*}
where $p_i^{*}$ is the reference position of defender $i$,  $q_i$ is the displacement from $p_1^{*}$ to $p_i^{*}$, and $\psi_i = \varphi + \frac{\pi}{2} + \frac{(2i-n-2)\beta}{2(n-1)}$ represents the orientation from $q_{i-1}$ to $q_i$. Five adjustable parameters in $\theta$, $p_{\rm c}\in\mathbb{R}^2$ $\varphi\in\mathbb{R}$, $\zeta\in\mathbb{R}$, and $\beta\in\mathbb{R}$ respectively correspond to the center, direction, spacing and opening angle of the arc formation pattern, which can realize the translation, rotation, scaling and deformation manipulation of the formation, respectively. Fig.~\ref{Fig_parametric_formation} gives a graphical explanation of the parameterized formation representation.

The arc formation can be maintained as each defender moves to its reference position
using the following control law:
\begin{equation}
    u_i = k_{\rm p}(p_i^{*}-p_i)+\dot{p}_{\rm c} + \frac{v_i}{C_{\rm d}}
    \label{Equ_control_law}
\end{equation}
where $k_{\rm p}$ is a constant. 

With this parametric representation, the pursuit strategy can be formulated as an update policy of the shape parameter $\theta$:
\begin{equation*}
    \dot{\theta} = \pi(\theta, s_{\rm a})
    \label{strategy_centralized}
\end{equation*}
where $s_{\rm a} = [p_{\rm a}^\top\ v_{\rm a}^\top]^\top$ is the state of the attacker.

\subsection{Distributed Negotiation on Shape Parameter}
The strategy presented in~\eqref{strategy_centralized} is centralized as it requires global state $\theta$ and global action $\dot{\theta}$. In distributed systems, a defender can only update its own estimated shape parameter based on local information. Subsequently, a negotiation among neighbors is necessary to maintain a consensus on the formation shape:
\begin{equation}
    \dot{\theta}_i = \pi(\theta_i,s_{\rm a}) + c_{\rm neg}\sum_{j\in\mathcal{N}_i}(\theta_j - \theta_i)
    \label{Equ_negotiation_para}
\end{equation}
where $\theta_i$ and $\dot{\theta}_i$ represent the estimated shape parameter of defender $i$ and corresponding update rate it takes, and $c_{\rm neg}$ is a constant coefficient. The first term of~\eqref{Equ_negotiation_para}, $\pi(\theta_i,s_{\rm a})$, indicates that each defender makes a decision based its own estimated shape parameter $\theta_i$ and the attacker state $s_{\rm a}$, and the last item is for consistency. The consensus of formation shape is guaranteed if the policy $\pi$ guides $\theta_i$ approaching instead of stay away from the optimal shape parameter $\theta^*$, that is,
\begin{equation}
    (\theta_i-\theta^*)^\top \pi(\theta_i,s_{\rm a})\leq 0.
    \label{convergence_condition}
\end{equation}
Here, if and only if $\theta_i=\theta^*$, $(\theta_i-\theta^*)^\top \pi(\theta_i,s_{\rm a})= 0$.

\begin{figure}
    \centering
    \includegraphics[width=0.8\linewidth]{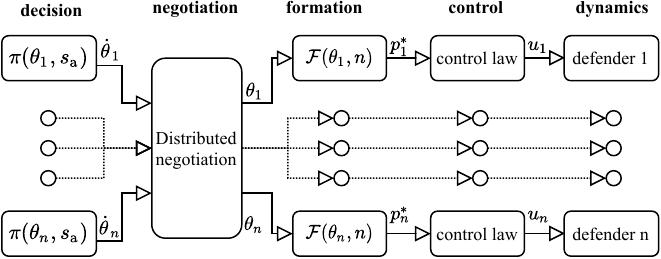}
    \caption{The distributed implementation workflow for defenders.}
    \label{Fig_alogrithm_flow}
\end{figure}

\begin{theorem}[convergence of policy negotiation]\label{theo_converge}
    If the condition~\eqref{convergence_condition} is satisfied, the negotiation strategy in~\eqref{Equ_negotiation_para} can guarantee that the estimated shape parameters of all defenders can converge to the optimal value $\theta^*$, that is, 
    \begin{equation*}
        \lim_{t\rightarrow\infty}\theta_i=\dots=\theta_n=\theta^*.
    \end{equation*}
\end{theorem}
\begin{IEEEproof}
    Define a Lyapunov function $V=\frac{1}{2}\sum_{i\in\mathcal{V}}\Vert \theta_i - \theta^*\Vert$. Its time derivative is $\dot{V}=\sum_{i\in\mathcal{V}} (\frac{\partial V}{\partial \theta_i})^\top \dot{\theta_i}$ with $\frac{\partial V}{\partial \theta}=\theta_i - \theta^*$. Considering~\eqref{Equ_negotiation_para}, we get 
    \begin{equation*}
        \begin{split}
            \dot{V} = &\sum_{i\in\mathcal{V}}(\theta_i-\theta^*)^\top\pi(\theta_i,s_{\rm a}) + \theta^*\sum_{i\in\mathcal{V}}\sum_{i\in\mathcal{N}_i}(\theta_j-\theta_i) \\
            &+ c_{\rm neg}\sum_{i\in\mathcal{V}}\sum_{i\in\mathcal{N}_i}\theta_i^\top(\theta_j-\theta_i)
            \label{Equ_laypunov}
        \end{split}.
    \end{equation*}
    According to~\eqref{convergence_condition}, the first term is no less than zero. The second term $\theta^*\sum_{i\in\mathcal{V}}\sum_{i\in\mathcal{N}_i}(\theta_j-\theta_i)=0$ as the topology is an undirected graph. The last term satisfies $c_{\rm neg}\sum_{i\in\mathcal{V}}\sum_{i\in\mathcal{N}_i}\theta_i^\top(\theta_j-\theta_i)=-c_{\rm neg}\sum_{(i,j)\in\mathcal{E}}(\theta_j-\theta_i)^\top(\theta_j-\theta_i)\leq 0$. In summary, $\dot{V}\leq 0$, and if and only if $\theta_i=\dots=\theta^*,\dot{V}=0$. Thus, $\lim_{t\rightarrow\infty}\theta_i=\dots=\theta_n=\theta^*$.
\end{IEEEproof}

Theorem~\ref{theo_converge} reveals that the negotiation law in~\eqref{Equ_negotiation_para} can accelerate the convergence of shape parameters of defenders, since the last term of~\eqref{Equ_laypunov} less than zero. This approach ensures that an ordered formation is established before reaching the optimal shape, thereby standardizing decision-making across different robots with consistent inputs and outputs. Consequently, only a single policy $\pi$ needs to be formulated, which can be implemented either centrally or in a distributed manner. The distributed implementation workflow for defenders is illustrated in Fig.~\ref{Fig_alogrithm_flow}.

\subsection{Imitation Learning for Shape Parameter}
\begin{figure}
    \centering
    \includegraphics[width=\linewidth]{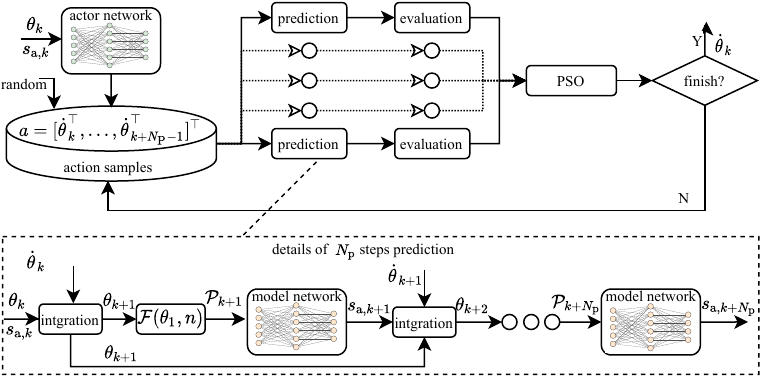}
    \caption{Algorithm flow for updating the shape parameter.}
    \label{Fig_algo_shape_parameter}
\end{figure}

The next work is to develop a shape parameter update strategy aimed at maximizing the defenders' ability to capture the attacker. 
To achieve this, we propose an imitation learning approach that integrates the rolling optimization of model predictive control. The algorithm flow is given in Fig.~\ref{Fig_algo_shape_parameter}.

The shape parameter update strategy presented in Fig.~\ref{Fig_alogrithm_flow}, denoted as $\pi(\theta_k,s_{{\rm a},k})$, receives the shape parameter $\theta_k$ and the attacker state $s_{{\rm a},k}$ at the current time (marked with subscript $k$), and output the shape parameter update rate $\dot{\theta}_k$. This process is equivalent to the rolling optimization process of model predictive control, and it plays the role of expert policy in the imitation learning perspective.
The distinctive features of this policy are that the initialization samples are provided by the actor network combined with the random strategy and the prediction model is learned using a model network. The details of these two networks and the initialization strategy will be introduced in Section~\ref{learning}.

\subsubsection{Expert policy}
\begin{figure}
    \centering
    \includegraphics[width=0.8\linewidth]{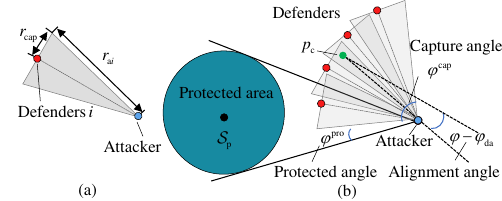}
    \caption{Graphical explanation of cost function design. (a) The capture angle of a single. (b) Schematic diagram of capture angle, protected angle, and alignment angle.}
    \label{Fig_cost}
\end{figure}

Let $N_{\rm c}$ and $N_{\rm p}$ be the number of control steps and prediction steps.
The predicted state trajectory consists of the next $N_{\rm p}$ states, denoted as $s = [s_{k+1}^\top,\dots,s_{k+N_{\rm p}}^\top]^\top$ with $s_{k} = [\theta_{k},s_{{\rm a},k}]$. The action sequence is denoted as $a = [\dot{\theta}_k^\top,\dots,\dot{\theta}_{k+N_{\rm p}-1}^\top]^\top$ with $\dot{\theta}_m = \dot{\theta}_k+N_{\rm c}-1,\forall m>k+N_{\rm c}-1$. The rolling optimization problem is described as 
\begin{equation}
    \begin{split}
        a^* =  &\mathop{\rm argmin}\limits_{a}:\sum_{m=0}^{N_{\rm p}-1} f(a_{k+m},s_{k+m+1})\\ 
        {\rm s.t.}\ & a_{\min} \leq a \leq a_{\max}
    \end{split}
    \label{Equ_optimization}
\end{equation}
where $a^{*}$ is the optimal action sequence, and $a_{\min}$ and $a_{\min}$ are the minimum and maximum bounds of $a$ depended on the mobility of the defenders. The function $f(a_{k+m},s_{k+m+1})$ represents the cost of each step, designed as
\begin{equation*}
    \begin{split}
        f = &-k^{\rm cap}(\varphi^{\rm cap})^2 + k^{\rm pro}(\varphi^{\rm pro})^2 + k^{\rm dis}\biggl(\frac{1}{n}\sum_{i=1}^{n}r_{ai}\biggr)^2 \\
        &+ k^{\rm ali}(\varphi - \varphi_{\rm da})^2 + k^{\rm ene} a^\top a
    \end{split}.
    \label{Equ_cost_design}
\end{equation*}
Here, coefficients $k^{\rm cap}$, $k^{\rm pro}$, $k^{\rm dis}$, $k^{\rm ali}$, and $k^{\rm ene}$ are the coefficients corresponding to capture angle, protected angle, average distance, alignment angle, and energy, respectively. The capture angle $\varphi^{\rm cap}$ is the angle range at which defenders can capture the attacker, the protected angle $\varphi^{\rm pro}$ is the angle range at which the attacker can reach the protected area without obstruction, $r_{ai}$ is the distance from defender $i$ to the attacker, and $\varphi_{\rm da}$ is the orientation from the center of the defenders to the attacker. Fig.~\ref{Fig_cost} gives a graphical explanation of these variables, where $r_{\rm cap}$ represents the defender's capture distance on the attacker.

The optimization problem~\eqref{Equ_optimization} is solved using the particle swarm optimization (PSO) algorithm.
Initialize a batch of action sequence $a_m^0,m=1,2,\dots$, each with a random initial change rate $\dot{a}_m^0$, where the subscript represents the sample label and the superscript represents the number of iterations. Let $a^{k}_{\rm gbest}$ and $a^{k}_{\rm pbest}$ be the global best and partial best samples from the first $k$ iterations, respectively. Then the update rules for each sample are as follows:
\begin{equation*}
    a_m^{k+_1} = a_{m}^{k} + \dot{a}_m^{k+1}
\end{equation*}
with 
\begin{equation*}
    \dot{a}_m^{k+1} = \omega\dot{a}_m^k + c_1r_1(a^{k}_{\rm gbest}-a_{m}^{k}) + c_2r_2(a^{k}_{\rm pbest}-a_{m}^{k})
\end{equation*}
where $\omega$ is the inertia weight, $c_1$ and $c_2$ are two constant weights, and $r_1,r_2\in[0,1]$ are two random weights. After a given number of iterations, the first term of the global best action sequence $\dot{\theta}$ is used to update the formation parameter.

\subsubsection{Imitation learning}\label{learning}

Due to limited computing resources and real-time requirements, the solution of~\eqref{Equ_optimization} obtained using the PSO algorithm is a suboptimal one. Its performance is significantly influenced by the accuracy of the prediction model and quality of the initial samples. To this end, this paper employs an imitation learning technique that leverages data generated from interactions with the environment to continuously enhance both aspects. Specifically, a model network is used to fit the state transitions of attackers in the actual environment, and an actor network is employed to imitate the historical outputs of the PSO algorithm.

To predict the state transition of the attacker, we build a structured neural network based on existing cognitive knowledge, enabling rapid learning of the attacker's strategy with a limited number of samples. It is assumed that the dynamics of the attacker and its basic properties of approaching the protected area and evading the defender are known, while the parameters in the attacker's strategy are unknown. Thus, the structure of~\eqref{Equ_dynamics_attacker} and~\eqref{Equ_control_attacker} can be used to construct the network.

\begin{figure}
    \centering
    \includegraphics[width=0.35\textwidth]{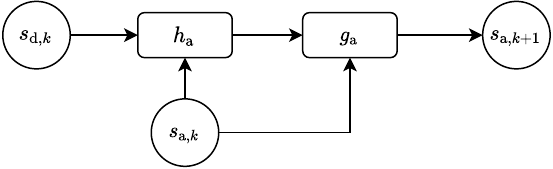}
    \caption{Structured network.}
    \label{Fig_structured_network}
\end{figure}

The dynamics~\eqref{Equ_dynamics_attacker} and strategy~\eqref{Equ_control_attacker} of the attacker can be respectively rewritten as 
\begin{equation}
    s_{{\rm a},k+1} = h_{\rm a}(s_{{\rm a},k},u_{{\rm a},k+1})
\end{equation}
and 
\begin{equation}
    u_{{\rm a},k+1} = g_{\rm a}(s_{{\rm a},k},s_{{\rm d},k}|\phi)
\end{equation}
Then the structured network can be formulated as
\begin{equation*}
    s_{{\rm a},k+1} = m(s_{{\rm a},k},s_{{\rm d},k}|\phi)=h_{\rm a}(s_{{\rm a},k},g_{\rm a}(s_{{\rm a},k},s_{{\rm d},k}|\phi))
\end{equation*}
with 
\begin{equation*}
    \phi = [ k_{\rm ap}, k_{\rm ad}, r_{\rm ad}^{\rm safe}, r_{\rm ad}^{\rm avo} ]^\top
\end{equation*}
where $s_{{\rm a},k}$ and $s_{{\rm d},k}$ are the state of the attacker and the defender closest to the attacker, and $\bm{\phi}$ is the trainable weights of the network. Since the center of the protection area can be set as the origin of the coordinate system, there is no need to include $p_{\rm p}$ in the network input. A graphical interpretation of the structured network $s_{{\rm a},k+1} = m(s_{{\rm a},k},s_{{\rm d},k}|\phi)$ is given in Fig.~\ref{Fig_structured_network}.

In each execution step, the current state $\bm{s}_{{\rm a},k}$ and $\bm{s}_{{\rm d},k}$ and the next state $\bm{s}_{{\rm a},k+1}$ can be collected. These data form a sample $\mathcal{B}_k^{m}=[\bm{s}_{{\rm a},k},\bm{s}_{{\rm d},k},\bm{s}_{{\rm a},k+1}]$ and are stored in a model replay buffer, denoted as $\mathcal{B}^{m}$. For each mini-batch sample $\mathcal{B}_{\rm mb}^{\rm m} \subseteq \mathcal{B}^{\rm m}$, the loss function is defined as 
\begin{equation*}
    l^{\rm m} = \sum_{\mathcal{B}_k^{\rm m}\in \mathcal{B}_{\rm mb}^{\rm m}} (\Delta\bm{s}_{k+1})^\top W^{\rm m} \Delta\bm{s}_{k+1}
\end{equation*}
with 
\begin{equation*}
    \Delta s_{k+1} = s_{{\rm a},k+1}-m(s_{{\rm a},k},s_{{\rm d},k})
\end{equation*}
where $W^{\rm m}$ is a $4\times 4$ diagonal matrix, with its diagonal elements corresponding to the weights of individual components.

Let $\alpha^{\rm m}$ be the actor learning rate and $\nabla_{\bm{\theta}} l^{\rm m}$ be the gradient of $l^{\rm m}$ with respect to $\phi$. The update law of $\phi$ is given by 

\begin{equation*}
    \phi_{t+1} = \phi_t - \alpha^{\rm m}\nabla_{\phi} l^{\rm m}.
\end{equation*}

The actor network learns the shape parameter policy by imitating the historical output of the PSO algorithm, denoted as $\pi^{\rm net}(\bm{\theta},\bm{s}_{\rm a}|\psi)$ where $\psi$ is the trainable weights. It employs a fully connected architecture with two hidden layers, each containing 18 nodes.
After each rolling optimization step, $[\bm{\theta}_k,\bm{s}_{{\rm a},k},\dot{\bm{\theta}}^{*}_k]$ is stored in a actor replay buffer, denoted as $\mathcal{B}^{\rm \pi}$. Here, the label $\dot{\bm{\theta}}^{*}$ is the output of PSO. For each mini-batch samples $\mathcal{B}^{\rm \pi}_{\rm mb} \subseteq \mathcal{B}^{\rm \pi}$, the loss function of strategy learning is 
\begin{equation*}
    l^{\pi} = \sum_{\mathcal{B}_k^{\rm \pi}\in \mathcal{B}_{\rm mb}^{\rm \pi}} (\Delta \dot{\theta}_k)^\top W^{\rm \pi} \Delta \dot{\theta}_k
\end{equation*}
with 
\begin{equation*}
    \Delta \dot{\theta}_k = \dot{\theta}^{*}_k - \pi^{\rm net}(\theta_k,\bm{s}_{{\rm a},k})
\end{equation*}
where $W^{\rm \pi}$ is a diagonal matrix with dimensions matching those of $\theta$, whose diagonal elements correspond to the weights of individual components.

\begin{table}
    \centering
    \caption{Parameter setting}
    \label{Table_sim_para}
    \begin{tabular}{cccccc}
        \toprule
        \centering
        Parameter & Value & Parameter & Value & Parameter & Value \\
        \midrule
        $u_{\rm d}^{\max}$ & $15{\rm m}/{\rm s}^2$ & $C_d$ & 7.5 & $r_{\rm cap}$ & 1m \\
        $u_{\rm a}^{\max}$ & $17.5{\rm m}/{\rm s}^2$ & $C_a$ & 7 & $r_{\rm ad}^{\rm safe}$ & 1m \\
        $v_{\rm d}^{\max}$ & 2m/s & $v_{\rm a}^{\max}$ & 2.5m/s & $r_{\rm com}$ & 4m \\
        $\rho_{\rm p}$ & 2m & $r_{\rm ad}^{\rm avo}$ & 10m & $k_{\rm ad}$  & 8 \\
        $\alpha^{\pi}$ & 0.001  & $\alpha^{\rm m}$ & 0.01  & $N_{\rm p}$ & 5 \\
        $c_1$ & 0.1  & $c_2$ & 0.1  & $N_{\rm c}$ & 2 \\
        \bottomrule
     \end{tabular}
     \vspace{-1em}
\end{table}

\begin{table}
    \centering
    \caption{Monte Carlo simulation results of the actor network and expert policy}
    \label{Tab_compare}
    \begin{tabular}{cccc}
        \toprule
        Method & Total cases  & Success rate & Average time  \\
        \midrule
        Actor network & 1000  & 0.401 & 54.5s  \\
        expert policy & 1000  & 0.993 & 10.3s  \\
        \bottomrule
    \end{tabular}
\end{table}

Let $\alpha^{\rm \pi}$ be the model learning rate and $\nabla_{\bm{\psi}} l^{\rm \pi}$ be the gradient of $l^{\rm \pi}$ with respect to $\psi$. The update law of $\psi$ is given by 
\begin{equation*}
    \psi_{t+1} = \psi_t - \alpha^{\pi}\nabla_{\psi} l^{\rm \pi}.
\end{equation*}

The trained action network $\pi^{\rm net}(\bm{\theta},\bm{s}_{\rm a})$, in addition to serving as a separated strategy for updating shape parameters, can also be integrated into the initialization strategy of rolling optimization problems to provide better initial actions. 
The specific initialization strategy is as follows.

Let $N$ be the number of initial samples. The first and second samples directly use the output of $\pi^{\rm net}(\bm{\theta},\bm{s}_{\rm a})$ and the one of the last step, respectively; the next $N/2-1$ samples are generated by sampling from a normal distribution $N(\mu,\sigma^2)$ with $\mu=\pi^{\rm net}(\bm{\theta},\bm{s}_{\rm a})$ and $\sigma^2$ is a given constant; the remaining sample are obtained through a uniform distribution. 
As the network output gradually approaches the optimal solution, the PSO algorithm can yield improved action sequences. 
This, in turn, contributes to the ongoing enhancement of the network's performance. 
From the perspective of imitation learning, the PSO-based rolling optimization process plays the role of an expert policy. The actor network and model network can improve the initial action sequences and prediction accuracy, enabling the learned actor network to achieve superior performance than the expert policy. In addition, performance can be further improved through data augmentation. By randomly generating virtual shape parameters and attacker states and then obtaining the corresponding solutions, a variety of training samples independent of the actual motion trajectory can be collected and used in the training of the action network.

\section{Simulation Results}\label{Sec_simulations}

In this section, we first present the simulation settings and learning process of the proposed scheme. Then, we provide two simulation examples to demonstrate the performance of our proposed scheme. 

\subsection{Simulation Setting and Learning Process}

The proposed strategy is trained in a simulated environment. The field size is 20m by 30m, with the protection area in the center. Defenders are randomly placed near the protection zone, while the attacker is randomly generated close to the boundary. To enhance computational efficiency, the number of samples per generation and the maximum number of iterations are both set to 5 when solving the rolling optimization problem using the PSO algorithm. Both the model network and the actor network are trained every 10 execution steps, with a mini-batch size of 16 during training. The remaining parameter settings are shown in Table~\ref{Table_sim_para}.

\begin{figure}[!t]
    \centering
    \includegraphics[width=\linewidth]{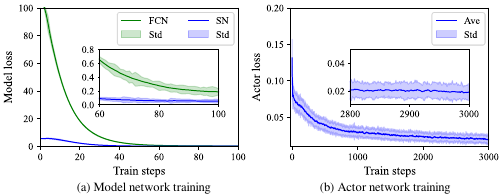}
    \caption{Curves of training loss. The solid curve represents the average value of the sample loss, and the width of the semi-transparent area represents the standard deviation of the sample loss distribution.}
    \label{Fig_train_loss}
\end{figure}

The loss curves of the model network and action network during training are shown in Fig.~\ref{Fig_train_loss}. For the model network training, in addition to the designed structured network (SN), a fully connected network (FCN) is also used for comparison. Fig.~\ref{Fig_train_loss}(a) shows that the structured network only needs dozens of steps to converge to a better value than the traditional fully connected network. Fig.~\ref{Fig_train_loss}(b) demonstrates the loss of the actor network convergences to below 0.02 in about 2000 steps.

\begin{figure}[!t]
    \centering
    \includegraphics[width=\linewidth]{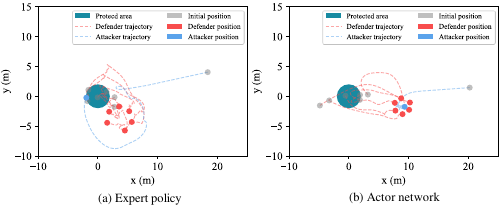}
    \caption{Simulation examples of expert policy and actor network.}
    \label{Fig_compare}
\end{figure}

\begin{figure}[!t]
    \centering
    \includegraphics[width=\linewidth]{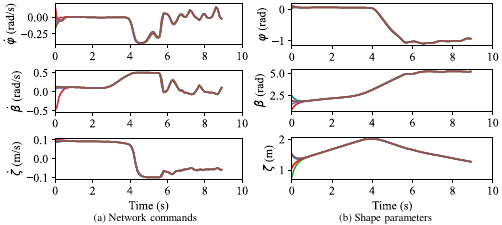}
    \caption{Curves of shape parameters and corresponding actor network commands during simulation. Each curve corresponds to a robot.}
    \label{Fig_compare_alpha}
\end{figure}

\subsection{Comparison Simulation}

This subsection compares the performance of the learned actor network with the expert policy through Monte Carlo simulation with random initial positions. The statistical results are shown in Table~\ref{Tab_compare}. We perform 1000 random simulations. 
The results indicate that the task success rate of the expert policy without learning is 0.401, with an average completion time of 54.5 seconds. In contrast, the actor network achieves a task success rate of 0.993 and an average completion time of 10.3 seconds. Due to inaccuracies in the prediction model and suboptimal optimization, the actions generated by the expert policy can result in the attacker escaping if the action magnitude is too large or causing a stalemate if the action magnitude is too small, which significantly prolongs the task completion time.

\begin{figure}
    \centering
    \includegraphics[width=0.48\textwidth]{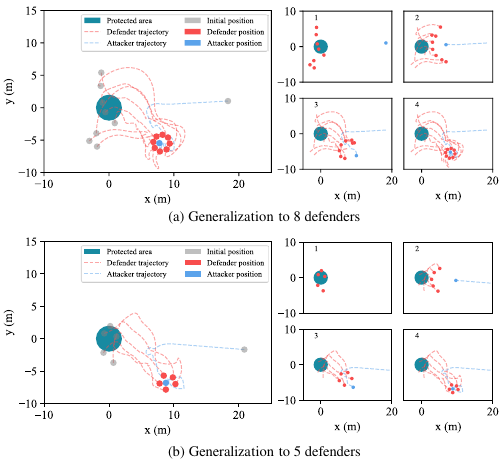}
    \caption{Quantitative generalization simulation}
    \label{Fig_number}
\end{figure}

Fig.~\ref{Fig_compare} gives two simulation examples using expert policy and actor network for shape parameter decisions. The motion trajectory in Fig.~\ref{Fig_compare} (a) shows that the attacker escapes from the defenders governed by the expert policy and enters the protected area. This is mainly attributed to the insufficient ability of the expert policy as well as the superior maneuvering ability of the attacker. Fig.~\ref{Fig_compare} (b) verifies that the learned actor network can successfully capture the attacker with more reasonable shape parameter updates.

Fig.~\ref{Fig_compare_alpha} shows the network putout commands and the shape parameters during the pursuit simulation using actor network in Fig.~\ref{Fig_compare} (b). The results demonstrate that the network commands and estimated shape parameters of each robot can quickly converge to the same value. This ensures that the defenders can form an ordered distribution of positions to collaboratively limit the motion of the attacker and ultimately capture the attacker, even though the decision-making and control process are implemented in a distributed manner. During the pursuit, the deformation parameter $\beta$ gradually decreases to complete the encirclement of the attacker. The scaling parameter $\zeta$, on the other hand, initially increases to prevent the attacker from entering the protected area and then decreases to achieve the capture of the attacker.

\subsection{Generalization Simulation}

To validate the proposed algorithm's generalization capability with respect to the number of defenders, we directly applied the strategy trained with six defenders to multi-robot pursuit tasks involving eight and five defenders. Each defender used the pre-trained network to make decisions regarding the shape parameters without requiring any modification to the network structure or retraining. 

In Fig.\ref{Fig_number}, the larger subplot on the left displays the trajectories of both parties during the simulation, while the smaller subplot on the right depicts snapshots at four different moments during this process. The results indicate that the actor network trained with six defenders can successfully coordinate eight and five defenders to complete the encirclement and capture of the attacker.

\section{Experiments}\label{Sec_experiment}

To validate the effectiveness of our proposed scheme beyond simulation, we implement the experiment demonstration using a group of 6 ground robots. 
The details are as follows. 

\begin{figure}[!t]
    \centering
    \includegraphics[width=0.8\linewidth]{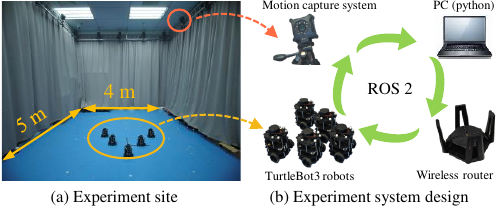}
    \caption{Experimental site and experimental system design}
    \label{Fig_experiment_setting}
\end{figure}

\subsection{Experiment Setting}

This section conducts experiments using TurtleBot3 robots to verify the performance of the proposed algorithm in real distributed systems. As shown in~\ref{Fig_experiment_setting}, the size of the experiment site is 4m by 5m and it is equipped with a motion capture system to acquire the locations and orientations of the robots in real-time. Similar to~\cite{sun2023mean} and~\cite{mulroy2022using}, the proposed algorithm is deployed on a personal laptop, running multiple threads where each thread represents a robot. Only robots that are neighbors can exchange information with each other. Since the dynamics of the  TurtleBot3 robot is nonholonomic, we refer to the results in~\cite{zhao2017general} to convert the control input in~\eqref{Equ_control_law} to linear velocity $v_i$ and angular velocity $\omega_i$. These control commands are then distributed to robots using the ROS2 network at a frequency of 15 Hz. The maximum linear and angular velocity of the defenders are set to $v_{\rm d}^{\rm max}$=0.1m/s and $\omega_{\rm d}^{\rm max}$=1.2rad/s, respectively, while those for the attacker are set to $v_{\rm a}^{\rm max}$=0.15m/s and $\omega_{\rm a}^{\rm max}$=2.4rad/s, respectively. The remaining parameters are set to be the same as those used in the simulation.

\subsection{Shape Manipulation Experiment}

\begin{figure}
    \centering
    \includegraphics[width=\linewidth]{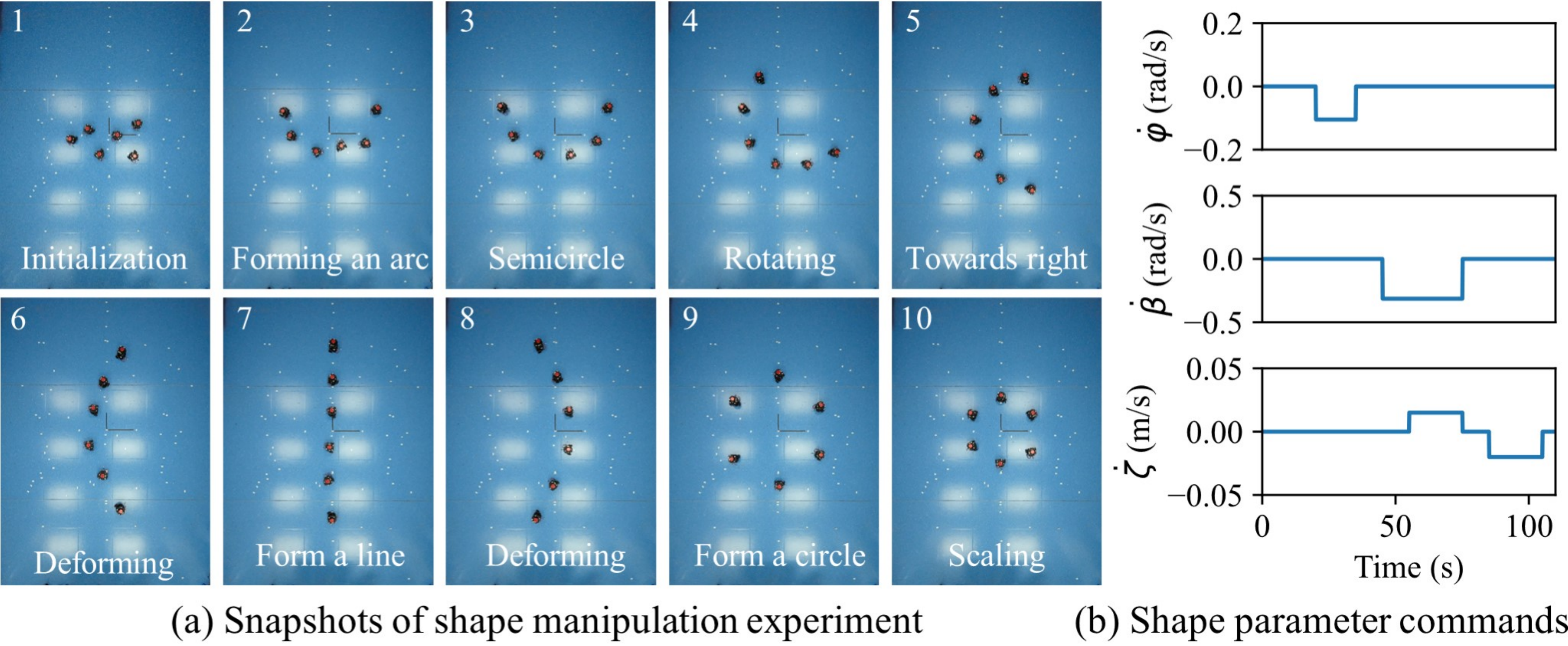}
    \caption{Results of shape manipulation experiment.}
    \label{Fig_manu}
\end{figure}

The first experiment, referred to as the shape manipulation experiment, demonstrates the algorithm's capability for flexible manipulation of positional distributions, which provides significant potential for defenders to collaboratively capture high-speed attackers. In this experiment, the shape parameter update rate commands are distributed uniformly by a central node. As illustrated in Fig.~\ref{Fig_manu}, by simply adjusting a few shape parameters, the defenders can exhibit various continuous shape transformations, including rotation, deformation, and scaling.

\subsection{Multi-Robot Pursuit Experiment}

\begin{figure}
    \centering
    \includegraphics[width=\linewidth]{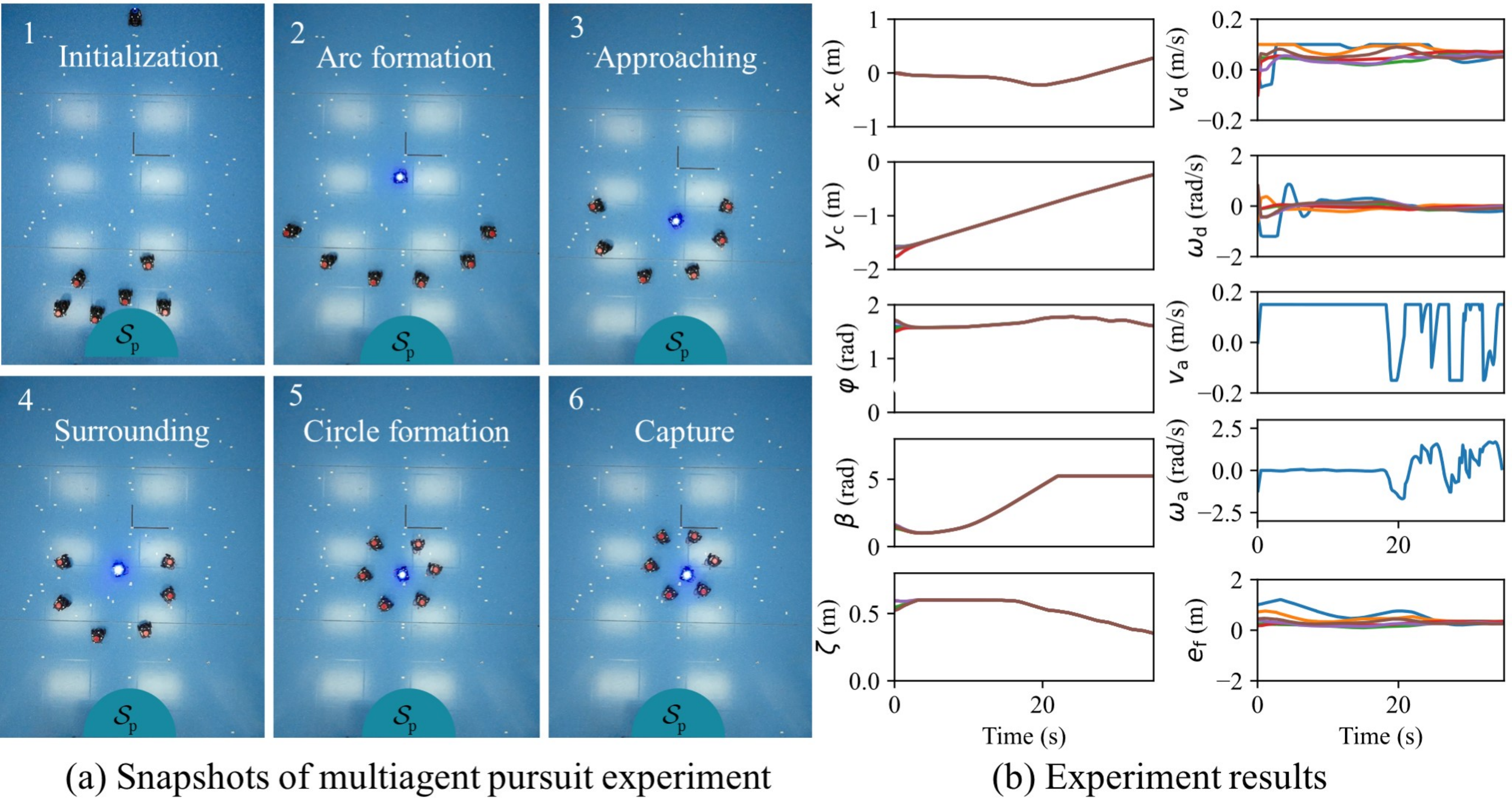}
    \caption{Results of multi-robot pursuit experiment.}
    \label{Fig_pursuit}
\end{figure}

The second experiment is a multi-robot pursuit task involving six defenders and one attacker equipped with a blue LED light. Each defender utilizes a pre-trained actor network to make decisions based on its local information. As shown in Fig.~\ref{Fig_pursuit} (a), the defenders quickly form an orderly formation and, upon approaching the attacker, adjust the deformation parameter $\beta$ to encircle the attacker. Finally, they successfully capture the attacker by reducing the scaling parameters. The first column of Fig.~\ref{Fig_pursuit} (b) presents the shape parameters estimated by each defender throughout the experiment. It can be observed that these parameters quickly converge from randomly initialized values. The second column shows the linear and angular velocities of both the defenders and the attacker, as well as the formation maintenance error of the defenders. Despite the attacker having higher angular and linear velocities and continuously struggling after being surrounded, it ultimately cannot avoid being captured by the defenders.

\section{Conclusion}\label{Sec_conclusion}

This paper proposes a multi-robot pursuit method using parameterized formations. Defending robots achieve the ability to capture the superior attacker by forming a parameterized formation that can be flexibly manipulated through five adjustable parameters. 
Moreover, we present an imitation-learning based approach combined with model predictive control for learning the shape parameter update strategy, endowing defending robots with the capability of distributed autonomous decision-making. Numerical simulations and physical experiments both demonstrate that, with the proposed method, defending roots can quickly learn effective strategies to capture the attacker, and the learned strategies can be directly applied to tasks with different numbers of defending robots. 

\bibliographystyle{IEEEtran}
\bibliography{reference}

\end{document}